\documentclass[conference]{IEEEtran}
\IEEEoverridecommandlockouts
\usepackage{cite}
\usepackage{amsmath,amssymb,amsfonts}
\usepackage{algorithmic}
\usepackage{graphicx}
\usepackage{textcomp}
\usepackage{xcolor}
\usepackage{multirow}
\usepackage{amsmath}
\usepackage{cases}
\usepackage{url}
\def\BibTeX{{\rm B\kern-.05em{\sc i\kern-.025em b}\kern-.08em
    T\kern-.1667em\lower.7ex\hbox{E}\kern-.125emX}}
\begin{document}

\title{Scalable Image Coding for Humans and Machines\\
Using Feature Fusion Network\\

\thanks{The results of this research were obtained from the commissioned research (JPJ012368C05101) by National Institute of Information and Communications Technology (NICT), Japan.}
}
\makeatletter
\newcommand{\linebreakand}{%
  \end{@IEEEauthorhalign}
  \hfill\mbox{}\par
  \mbox{}\hfill\begin{@IEEEauthorhalign}
}

\makeatother
\author{\IEEEauthorblockN{Takahiro Shindo \qquad Taiju Watanabe \qquad Yui Tatsumi \qquad Hiroshi Watanabe}
\vspace{2mm}
\IEEEauthorblockA{\textit{Waseda University} \\
Tokyo, Japan \\
{\small\{taka\_s0265@ruri., \ lvpurin@fuji., \ yui.t@fuji., \ hiroshi.watanabe@\}waseda.jp}}
}

\maketitle

\begin{abstract}
As image recognition models become more prevalent, scalable coding methods for machines and humans gain more importance.
Applications of image recognition models include traffic monitoring and farm management.
In these use cases, the scalable coding method proves effective because the tasks require occasional image checking by humans.
Existing image compression methods for humans and machines meet these requirements to some extent.
However, these compression methods are effective solely for specific image recognition models.
We propose a learning-based scalable image coding method for humans and machines that is compatible with numerous image recognition models.
We combine an image compression model for machines with a compression model, providing additional information to facilitate image decoding for humans.
The features in these compression models are fused using a feature fusion network to achieve efficient image compression.
Our method's additional information compression model is adjusted to reduce the number of parameters by enabling combinations of features of different sizes in the feature fusion network.
Our approach confirms that the feature fusion network efficiently combines image compression models while reducing the number of parameters. 
Furthermore, we demonstrate the effectiveness of the proposed scalable coding method by evaluating the image compression performance in terms of decoded image quality and bitrate.
Code is available at {\color{magenta}\textit{\url{https://github.com/final-0/ICM-v1}}}.
\end{abstract}

\begin{IEEEkeywords}
Scalable Image Coding, Image Coding for Machines, Learned Image Compression
\end{IEEEkeywords}

\section{Introduction}
Scalable coding aims to decode images for multiple purposes.
A well-known scalable coding method decodes a low-resolution image from minimal information. 
Additionally, a high-resolution image can be decoded by supplying extra information to this coding method \cite{e1,e2}.
Hence, this method provides two types of decoded images suitable for human vision.
While there is a demand for image coding for human vision, there is also interest in image coding for image recognition models. 
This has led to the development of the research fields known as Video Coding for Machines (VCM)\cite{a1,a2,a3,d4,d5,f1,f2} and Image Coding for Machines (ICM)\cite{a4,a5,a6,d3,a7}, which are standardized by MPEG and JPEG.
Therefore, scalable coding methods, which provide step-by-step image compression methods for human and machine, are attracting attention \cite{b1,b2,b3,b4,b5,d1,d2,d7,d9}.
H. Choi's redearch infers that the amount of information in an image required for image recognition is less than or part of that required for human vision \cite{b5}.
Therefore, efficient image compression can be accomplished by applying decoded images for image recognition model to reconstruct images for human vision.

Scalable coding methods that hierarchically implement image compression methods for both machines and humans are mainly needed for traffic and farm surveillance.
In these use cases, most images are analyzed by image recognition models, with occasional verification by the human eye.
Object detection models and segmentation models are employed to analyze the images.
These models estimate the location, size, and types of objects and background present in an image.
Therefore, it is necessary to have two different image compression methods: one for these image recognition models and the other for human checking.
The majority of the proposed scalable coding methods are optimized to increase the accuracy of a specific image recognition model. 
However, these methods may struggle when multiple image recognition tasks are required.
Therefore, it is necessary to consider scalable coding methods for any image recognition model to increase generality.
As Fig. \ref{fig:flow}(a) illustrates, when the image compression model for machines is optimized based on a particular image recognition model, it is necessary to decode the image for humans with a corresponding specific additional information compression model.

\begin{figure}[bt]
    \centerline{\includegraphics[width=0.98\columnwidth]{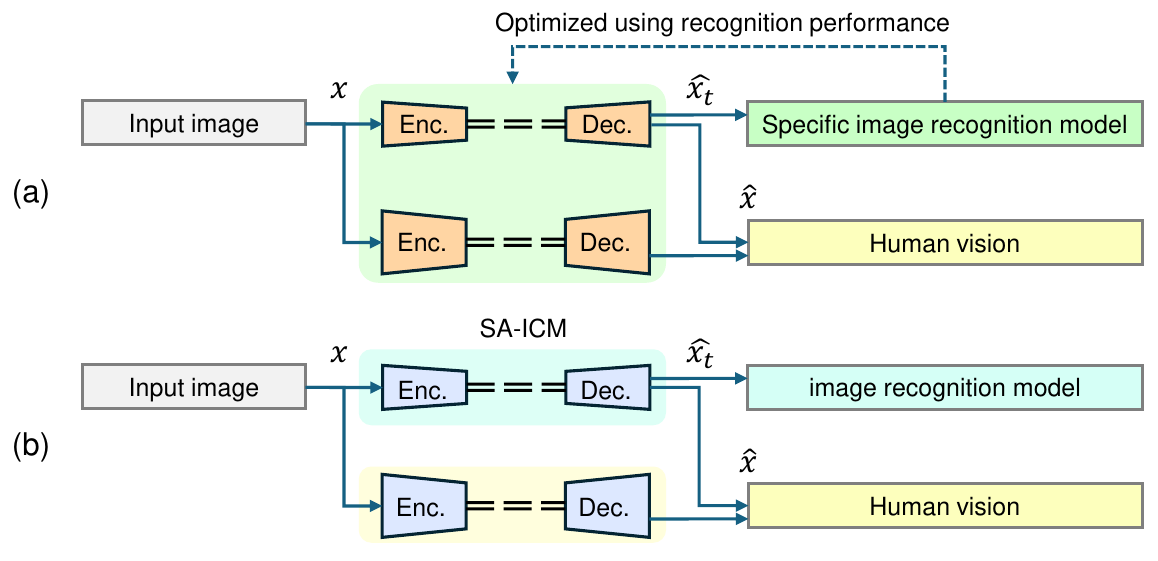}}
    \caption{Image processing flow of the scalable image coding method. (a): Conventional method aimed for a specific image recognition model. (b): Proposed method compatible for various image recognition models.}
    \label{fig:flow}
    \end{figure}

We propose a scalable coding method that incorporates a robust ICM method against changes in image recognition models.
This proposed method differs from conventional scalable coding methods by performing stepwise image decoding for human vision and any image recognition model.
With our proposed model, there is no need to modify the image compression model on the device whenever a different image recognition model is desired.
In this method, SA-ICM \cite{b6} is used as the ICM method.
As Fig. \ref{fig:flow}(b) indicates, images decoded in this way are suitable for a variety of image recognition tasks.
The difference between the SA-ICM decoded image and the original image is treated as additional information and compressed using Learned Image Compression (LIC)\cite{b7,b8,b9,d6,d8}.
Through the fusion of the additional information and SA-ICM features in the feature space, efficient image reconstruction for human vision can be achieved.

The remainder of this paper is organized as follows: Section 2 and Section 3 cover related work and the proposed method, respectively. 
Section 4 discusses the experiments and results, whereas the last section presents the conclusion.

\section{Related Work}

\subsection{Scalable Image Coding for Humans and Machines}
With the rapid expansion in the use of image recognition models, the demand for image compression methods to support them is increasing.
In particular, image analysis methods powered by image recognition models are expected to become popular due to their usefulness in monitoring people and traffic as well as managing farm animals.
On the other hand, human verification is sometimes necessary for monitoring purposes insted of solely relying on AI for image analysis.
To achieve both of these purposes, scalable image coding methods have been studied.
Since it is generally thought that the amount of information in an image required for image recognition is less than that needed for human, images for humans can be decoded by supplementing images decoded for machines with additional information.

Recently, H. Choi \textit{et al}. proposed a scalable coding method for object detection models, object segmentation models, and humans \cite{b5}.
In this study, scalable codecs are created by dividing the intermediate features of the image recognition model into two categories: features for image recognition and features for additional information.
The method achieves strong compression performance by leveraging the model's characteristics, though the compression method is optimized in that particular model.
Separately, A. Harell \textit{et al}. utilize VVC \cite{b10} to compress additional information \cite{b11}, known as VVC+M.
The difference between the decoded image for image recognition and the original image is encoded through VVC-inter compression to decode the image for human vision.
VVC is a rule-based algorithm and a video compression method that is robust to changes in the input image.
Therefore, VVC+M can be incorporated into any machine vision codec, enabling image decoding for humans and machines.
This method can be combined with a variety of ICM methods, yet it still leaves room for optimization through LIC.

\subsection{Channel-Conditional Learned Image Compression}
An LIC model is a neural network-based image compression model.
Most LIC models are composed of a main network and a hyperprior network proposed by J. Ballé \textit{et al}.\cite{c1,c2}.
The hyperprior model is a technique for capturing spatial dependency in latent representations and learning a probabilistic model for entropy coding.
The following loss function is used to train a general LIC model.
\begin{equation}
    \mathcal{L}=\mathcal{R}(y)+\mathcal{R}(z)+\lambda \cdot mse(x,\hat{x}).
  \end{equation}
In (1), $y$ is the encoder output of the LIC model, $\mathcal{R}(y)$ is the bitrate of $y$ and is calculated using compressAI \cite{c12}. 
$z$ is the hyperprior-encoder output, and $\mathcal{R}(z)$ is the bitrate of that feature. 
$x$ represents the input image, and $\hat{x}$ denotes the decoder output image.
$mse$ is the mean squared error function and $\lambda$ stands for a constant to control the rate.
The LIC model learns to decode the original image while dropping the amount of information in the image.

The Channel-Conditional LIC model is an image compression model proposed by D. Minnen \textit{et al}.\cite{c3}.
This model divides the output of the main network into multiple features and prepares a probabilistic model for coding each feature.
The input image $x$ is first converted into a feature $y$ by the main network on the encoder side.
The feature $y$ is divided into $n$ features $\{y_1,y_2, ..., y_n\}$, each of which is compressed.
The hyperprior network learns a probabilistic model to encode $y_1$, thereby generating $\hat{y_1}$. 
Then, $\hat{y_1}$ is used in a probabilistic model to encode $y_2$, creating $\hat{y_2}$. 
By repeating this process, we ultimately obtain $\hat{y}$, a combination of $\{\hat{y_1},\hat{y_2}, ..., \hat{y_n}\}$.
Therefore, with its step-by-step image compression method, the Channel-Conditional LIC model allows for efficient image compression.
\begin{figure}[bt]
    \centerline{\includegraphics[width=0.95\columnwidth]{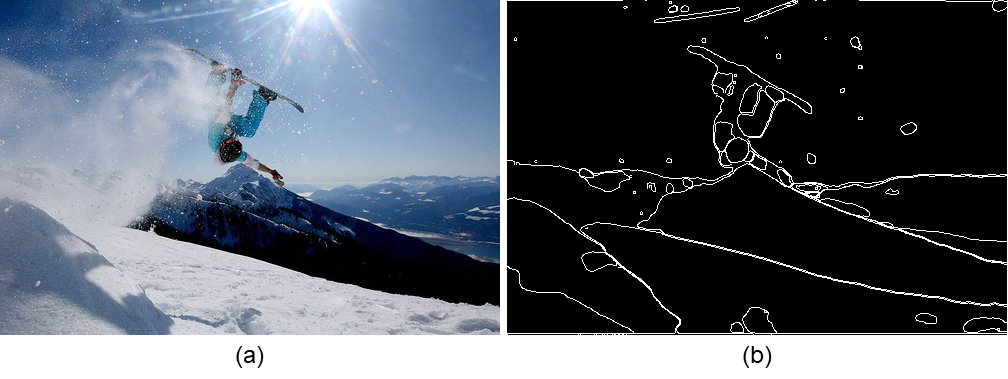}}
    \caption{Examples of the mask image. (a) : Original image in COCO dataset. (b) : Mask image generated using Segment Anything Model.}
    \label{fig:mask}
    \end{figure}

\begin{figure*}[bt]
    \centerline{\includegraphics[width=1.96\columnwidth]{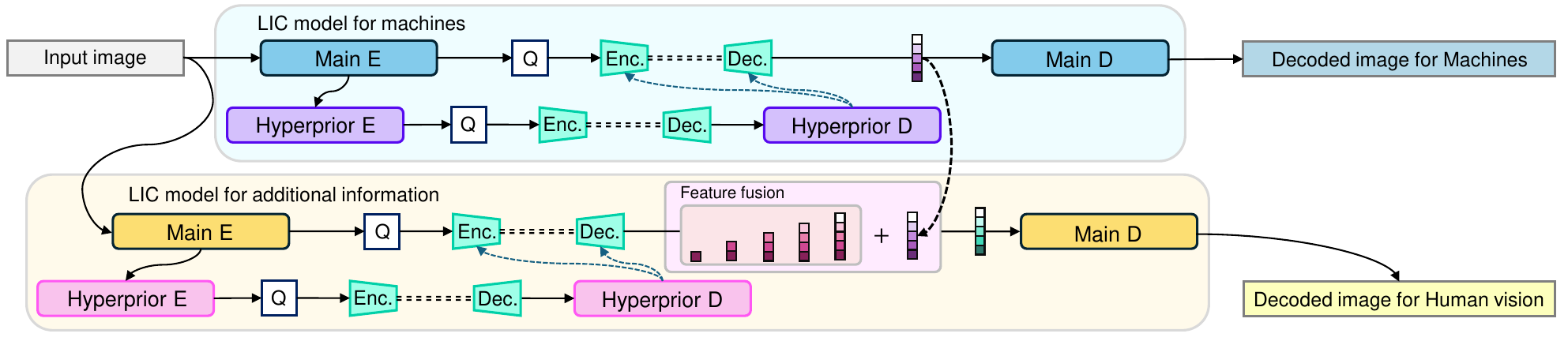}}
    \caption{The processing flow of the proposed scalable image coding method. The LIC model in the upper row is SA-ICM, which compresses images for machines. The LIC model in the lower is an additional information compression model, which converts coded images for machines into images for humans.}
    \label{fig:str}
    \end{figure*}
\begin{figure}[bt]
    \centerline{\includegraphics[width=0.96\columnwidth]{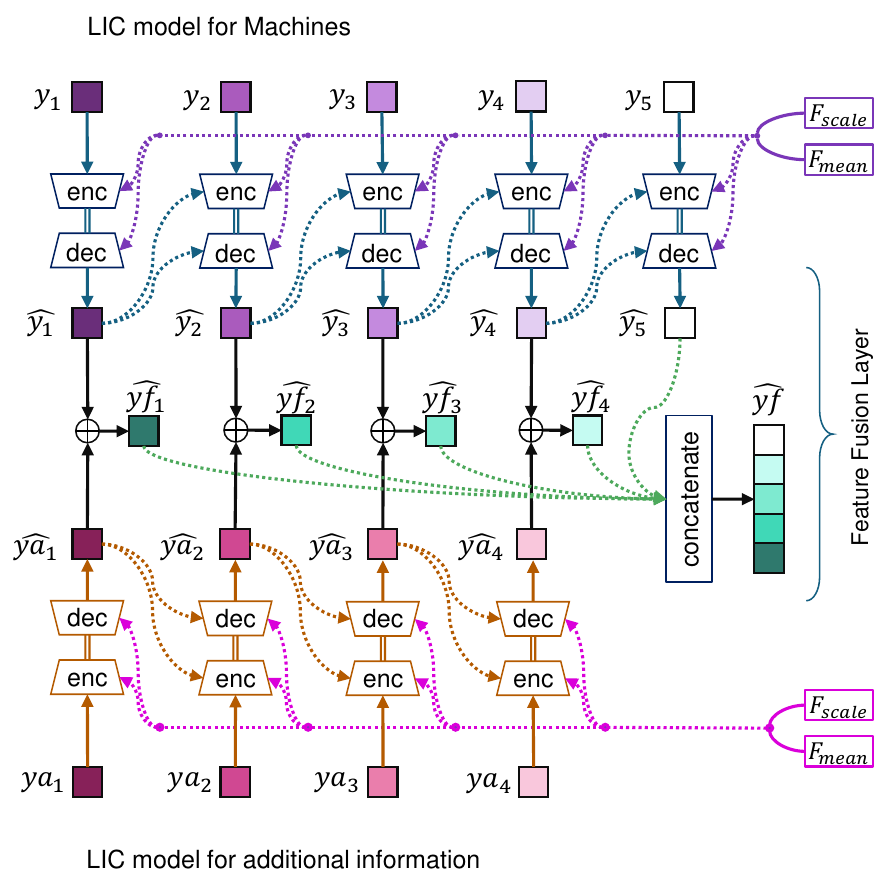}}
    \caption{Model structure of the feature fusion network. The purple and pink squares represent the features of the image compression model for machines and that of the additional information compression model, respectively.}
    \label{fig:ff}
    \end{figure}

\subsection{SA-ICM}
SA-ICM is an LIC model that learns only the edge information of images, with a model structure is identical to the LIC model proposed by J. Liu \textit{et al}.\cite{c4}.
By learning the edge information produced with Segment Anything \cite{c5}, we can create an LIC model capable of decoding only the edges of an image.
Meanwhile, the decoded image retains the color of the original image as well as the edge information of the object and background, making the image suitable for image recognition models.
This model has been proven to be an effective image compression method for object detection models\cite{c6}, object segmentation models\cite{c7}, and background segmentation models\cite{c8}.
The usefulness of SA-ICM is demonstrated using three different datasets\cite{c9,c10,c11}.
The loss function is expressed by the following equation:
\begin{equation}
    \mathcal{L}_{rl}=\mathcal{R}(y)+\mathcal{R}(z)+\lambda \cdot mse(x \odot m_{x},\hat{x} \odot m_{x}).
  \end{equation}
The symbols for variables and functions in (2) carry the same meaning as those in (1).
$m_{x}$ is the binary mask corresponding to $x$, created using Segment Anything.
An example of a mask image is shown in Fig. \ref{fig:mask}.
SA-ICM is a method to train an LIC model to encode and decode only the white areas of the mask image in Fig. \ref{fig:mask}.

\section{Proposed Method}

\begin{figure*}[bt]
    \centerline{\includegraphics[width=2.0\columnwidth]{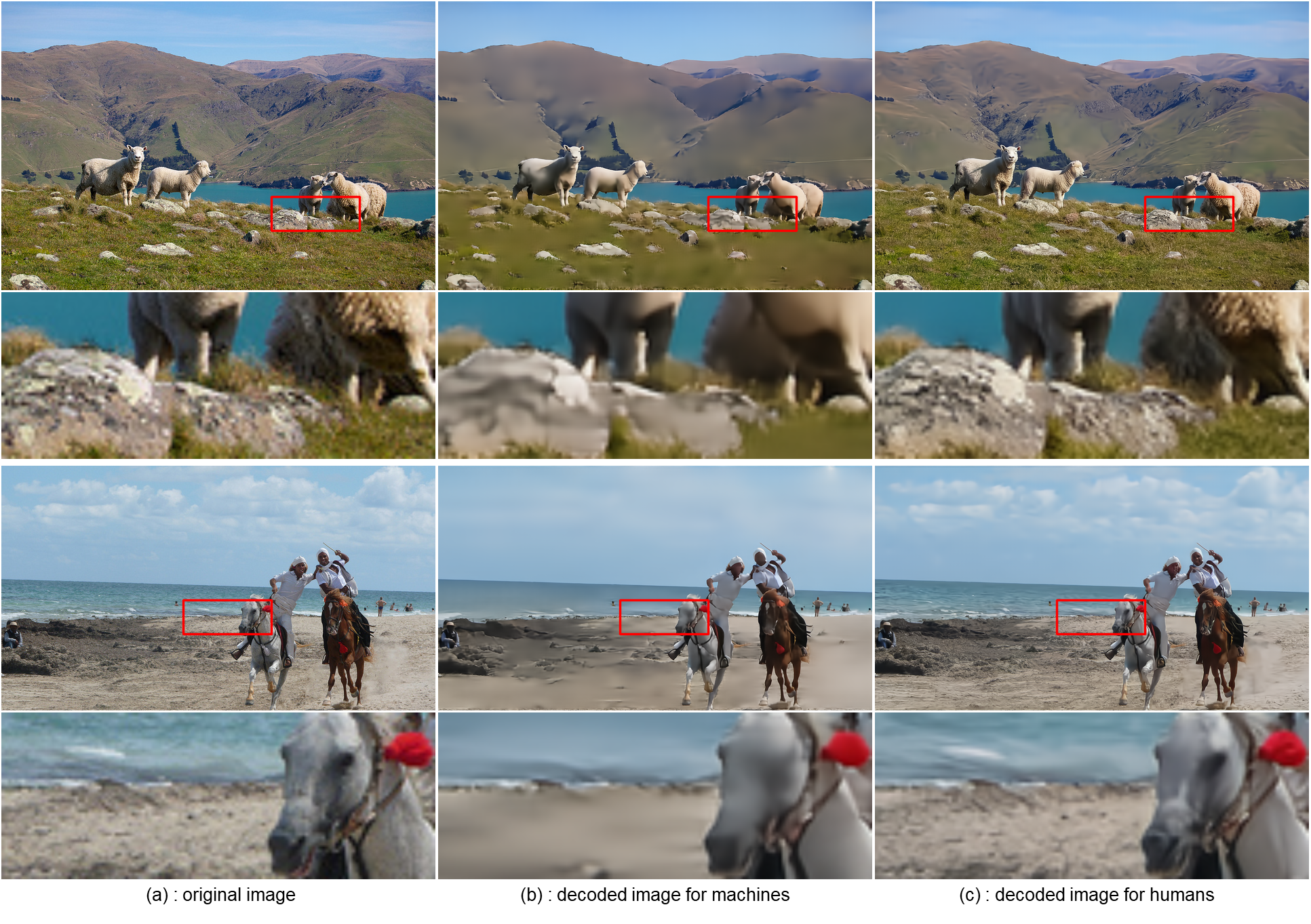}}
    \caption{Example of compressed images for humans and machines. (a) : original image. (b) : decoded images for machines using SA-ICM. (c) : decoded images for human vision using additional information.}
    \label{fig:coded}
    \end{figure*}
          
\subsection{Learning-based Scalable Image Coding}
We propose a scalable coding method that is robust to shifts in image recognition models.
Our method does not optimize a compression method exclusively for a specific image recognition model.
Thus, this method is applicable for compressing images across various image recognition models and human vision.
In other words, there is no need to modify the image compression method to accommodate a switch in an image recognition model.
Our method is implemented by connecting two LIC models.
The structure of the proposed model is shown in Fig. \ref{fig:str}.

In Fig. \ref{fig:str}, the upper network represents an image compression method designed for image recognition models, utilizing SA-ICM.
The network in the lower part of the proposed model is employed to compress additional information, converting a coded image for machines into an image for human vision.
In the feature space, the upper and lower networks are combined to efficiently reconstruct images adequate for humans.
Additional information compression model is trained using the following loss function:
\begin{equation}
    \mathcal{L}_{a}=\mathcal{R}(ya)+\mathcal{R}(za)+\lambda \cdot mse(x,\hat{x}).
  \end{equation}
In (3), $ya$ and $za$ denote the output of the encoder and hyper-encoder in the LIC model, respectively.
The symbols for the other variables and functions represent the same meaning as in (1).
Optimization using $mse$ allows the original image texture to be regained.

\subsection{Feature Fusion Network}
We apply a feature fusion network to combine the upper LIC model with the lower LIC model in Fig. \ref{fig:str}.
The structure of the feature fusion network is shown in the Fig. \ref{fig:ff}.
Both compression models are Channel-Conditional LIC models, and their decoder models receive segmented features.
SA-ICM applies the following equation to decode an image:
\begin{equation}
    \hat{x_t}=g_{machine}(\hat{y}),
  \end{equation}
\begin{equation}
    \hat{y} = conc(\hat{y_1},\hat{y_2}, ..., \hat{y_n}).
\end{equation}
In (4), $\hat{x_t}$ represents the decoded image for machines, $g_{machine}$ is the decoder function and $\hat{y}$ is the decoder input of the LIC model.
In (5), $\hat{y_i} (i=1,2,...,n)$ is the divided features and $conc$ stands for the concatenate function of these features.

The decoder of the additional information compression model receives additional information $\hat{ya_j}(j= 1,2,...,m)$ and the same $\hat{y_i} (i=1,2,...,n)$  as the SA-ICM decoder input.
By configuring the parameter $m$ to be less than the parameter $n$, the quantity of feature channels dedicated to representing supplementary information diminishes. 
These features are combined using a feature fusion network. 
To reduce the weight of additional information compression models, we implement a feature fusion network that allows the merge of LIC models with different feature sizes.
The feature fusion function for fusing features $\hat{y_i}$ and $\hat{ya_j}$ with different number of channels is shown below:
\begin{equation}
    \hat{yf_k} =
\left\{ 
\begin{array}{ll}
    \hat{y_k}+\hat{ya_k} & (1 \leq k \leq m) \\
    \hat{y_k}            & (m < k \leq n)
\end{array} \right.
\end{equation}
\begin{equation}
    \hat{yf} = conc(\hat{yf_1},\hat{yf_2}, ..., \hat{yf_n}).
\end{equation}
$\hat{yf_k} (k=1,2,...,n)$ is the feature input to the lower decoder found in Fig. \ref{fig:str}.
The remaining variables and functions align with those previously discussed.

The LIC model for additional information utilizes the feature $\hat{yf}$ to decode images targeted for humans.
This procedure is facilitated by the following equation:
\begin{equation}
    \hat{x}=g_{human}(\hat{yf}).
  \end{equation}
  In (8), $\hat{x}$ is the decoded image for humans, $g_{human}$ is the decoder function, and $\hat{yf}$ is the decoder input of the LIC model.

\section{Experiment}
\subsection{Performance in Image Compression for Humans}
We assess the image compression performance of the proposed scalable coding method.
Previously, we evaluated the compression performance for machines under the SA-ICM framework \cite{b6}. 
Therefore, we proceed to examine the compression performance for humans. 
The COCO-train dataset is used to train the model, and the COCO-val dataset is used for testing \cite{c9}. 
The LIC model is the model proposed by J. Liu \textit{et al}.\cite{c4}.
In this experiment, the parameter $n$ is fixed at 5, resulting in the split of feature $y$ into five distinct components within the image compression model for machines. 
Fig. \ref{fig:coded} illustrates how integrating additional information transforms the decoded image from SA-ICM into one closely resembling the input image's texture.

\begin{figure}[bt]
    \centerline{\includegraphics[width=1.0\columnwidth]{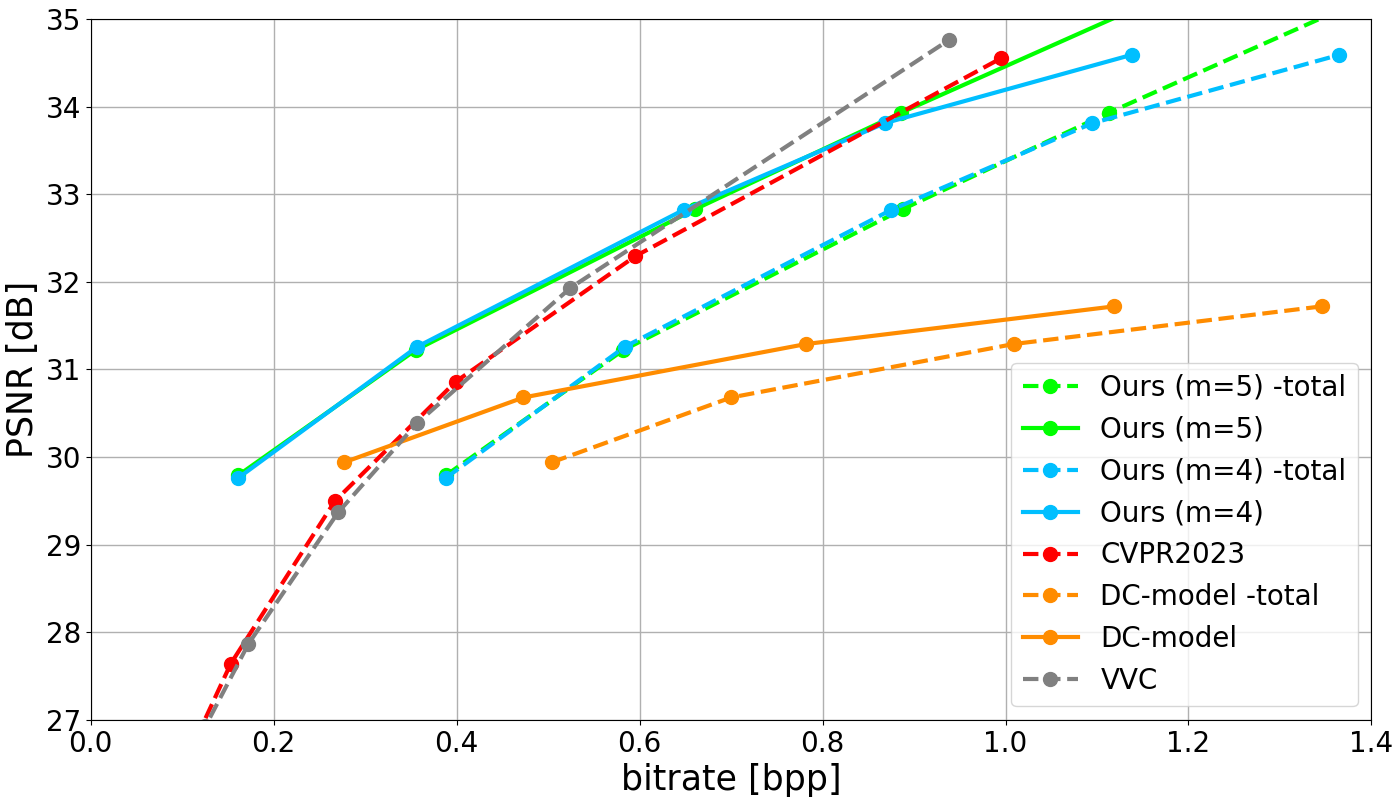}}
    \caption{Image compression performance of the proposed and comparative methods.}
    \label{fig:res1}
    \end{figure}
\begin{figure}[bt]
    \centerline{\includegraphics[width=1.0\columnwidth]{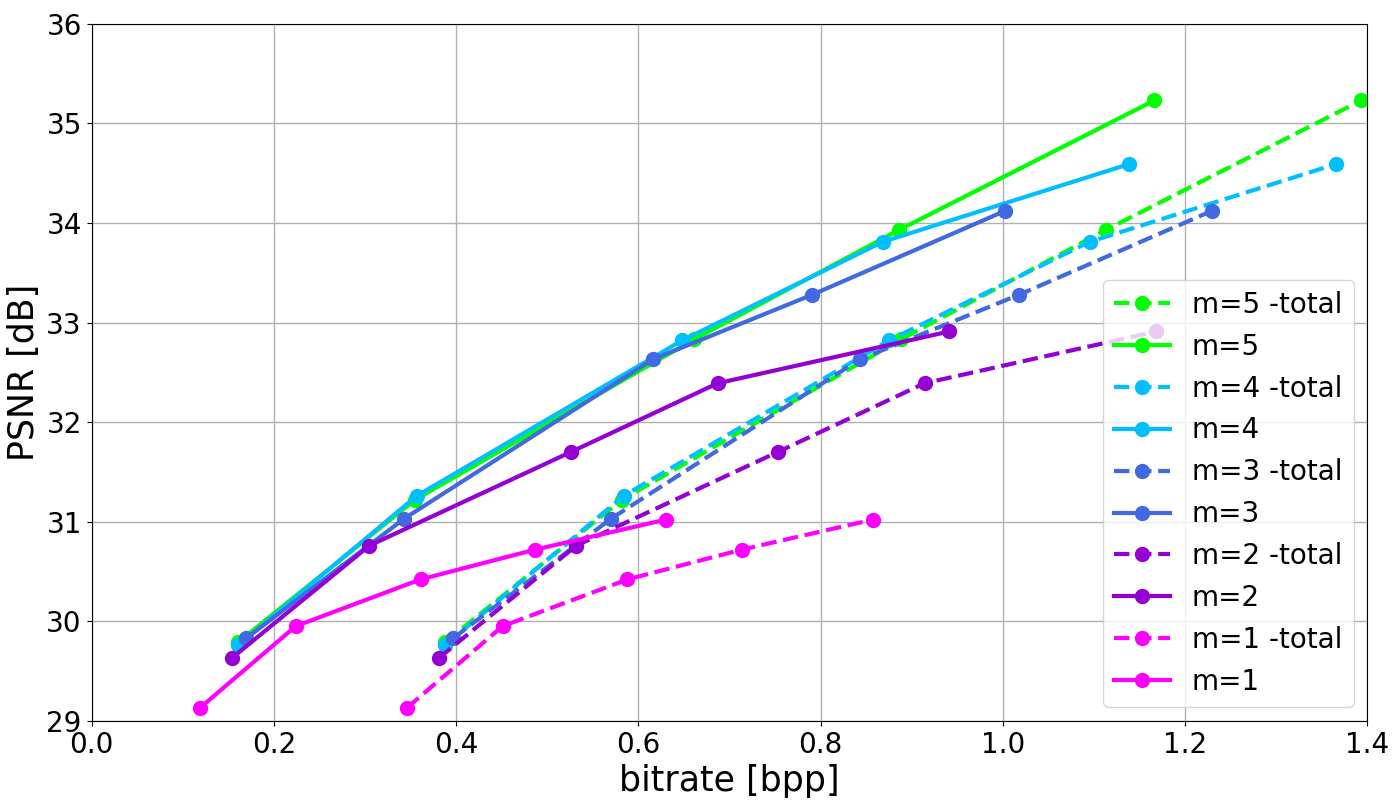}}
    \caption{Relationship between the change in parameter $m$ and compression performance of our proposed method.}
    \label{fig:res2}
    \end{figure}

The rate-distortion curve is drawn as shown in Fig. \ref{fig:res1}.
The light blue solid line illustrates the correlation between the quantity of supplementary information and image quality, while the light blue dotted line represents the relationship between the amount of additional information plus the image information for machines and image quality.
We compare the performance of the proposed method with the VVC\cite{b10} and the LIC model for human vision\cite{c4}.
The scalable coding method without feature fusion network is also used as a comparison method.
This method compresses the difference between the SA-ICM decoded image and the original image using LIC.
We call it the difference-compression model (DC model).
The orange curve represents the compression performance of the DC model.
As depicted in Fig. \ref{fig:res1}, our proposed method demonstrates superior performance compared to both the VVC and the LIC model for human, especially at lower bit rates.
In other words, the proposed method proves effective when the majority of images are incorporated into the image recognition model, with occasional human checks, as exemplified in the use case delineated in Section 1.
In addition, the superiority of the proposed method over the DC-model confirms the effectiveness of the feature fusion network. 

\subsection{Parameter Reduction in the Additional Information Compression Model}
As noted in Section 3.2, we investigate the impact of parameter reduction in the additional information compression model.
In this experiment, the parameter $n$ is fixed at 5, consequently implying that the value of $m$ falls within the range of greater than 1 and less than 5.
The size of the model can be adjusted based on the value of $m$.
Setting a larger value for $m$ increases the number of intermediate features.
Hence, the number of parameters also increases.
The relationship between the value of $m$ and the number of parameters of the additional information compression model is shown in Table \ref{tab:param}.
The relationship between the value of $m$ and the image quality of the decoded image is shown in Table \ref{tab:psnr} and Fig. \ref{fig:res2}.
These results indicate that decoding images for humans does not heavily rely on the value of $m$, particularly at low bitrates.
For instance, at bitrates around 0.6 [bpp], images of nearly identical quality are decoded whether $m$ is 3, 4, or 5.
Therefore, depending on the value of $m$, it is possible to reduce the number of parameters while maintaining the quality of the decoded image.
\begin{table}[t]
    \centering
    \caption{Relationship between the size of additional imfomation compression model and the value of $m$} \label{tab:param}
    \small
    \begin{tabular*}{8.8cm}{c|ccccc}
      \hline
      $m$ & 1 & 2 & 3 & 4 & 5\\  
  
      \hline
      \hline
      number of channels of $ya$ & 64 & 128 & 192 & 256 & 320\\
      number of params (M) & 45.9 & 51.7 & 58.7 & 67.0 & 76.6\\
      \hline
    \end{tabular*}
  \end{table}

\begin{table}[t]
\centering
\caption{Image quality (PSNR) in the decoded image for different values of $\lambda$ and $m$.} \label{tab:psnr}
\small
\begin{tabular*}{7.8cm}{p{1.5cm}|ccccc}
    \hline
    \multirow{2}{*}{\qquad$\lambda$} & \multicolumn{5}{c}{$m$} \\
                             & 1 & 2 & 3 & 4 & 5 \\
    \hline
    \hline
    \quad0.005  & 29.13 & 29.63 & 29.83 & 29.76 & 29.79\\
    \quad0.010  & 29.95 & 30.76 & 31.03 & 31.26 & 31.22\\
    \quad0.020  & 30.42 & 31.70 & 32.63 & 32.82 & 32.83\\
    \quad0.030  & 30.72 & 32.39 & 33.28 & 33.81 & 33.93\\
    \quad0.050  & 31.02 & 32.91 & 34.12 & 34.59 & 35.23\\
    \hline
\end{tabular*}
\end{table}
\section{Conclusion}
We propose a learning-based scalable image compression method for humans and machines.
Our method offers versatile image compression for various recognition models and can also decode images for humans by utilizing additional information.
To seamlessly integrate image data for machines with additional information, we introduce a feature fusion network and validate its efficacy through experiments.
Moreover, the feature fusion network enables the fusion of features of varying sizes and diminishes the parameter count in the additional information compression model.
We confirm the effectiveness of our method through the measurement of the additional information rate and the quality of decoded images.
On the other hand, in pursuit of enhanced coding performance, it is imperative to address the improvement in image compression methods for machines. 
Additionally, deliberations on further parameter reduction are warranted to alleviate the computational burden on edge devices and popularize the adoption of LIC-based image compression.

\vspace{12pt}
\end{document}